\documentclass[lettersize,journal]{IEEEtran}
\usepackage{amsmath,amsfonts}
\usepackage{algorithmic}
\usepackage{algorithm}
\usepackage{array}
\usepackage[caption=false,font=normalsize,labelfont=sf,textfont=sf]{subfig}
\usepackage[table]{xcolor}   
\definecolor{bestc}{HTML}{FF9999}
\definecolor{secondc}{HTML}{FFCC99}
\definecolor{thirdc}{HTML}{FFFF99}
\newcommand{\best}[1]{\cellcolor{bestc}#1}
\newcommand{\second}[1]{\cellcolor{secondc}#1}
\newcommand{\third}[1]{\cellcolor{thirdc}#1}
\usepackage{textcomp}
\usepackage{stfloats}
\usepackage{url}
\usepackage{verbatim}
\usepackage{graphicx}
\usepackage{cite}
\usepackage{booktabs}
\begin{document}

\title{PUGS: Physics-Guided Uncertainty Modeling for Underwater Dynamic 3D Gaussian Splatting}

\author{Tingjun Li,
          Yuxi~Wang$^{*}$,
          Guoqiang Zhong$^*$,~\IEEEmembership{Member,~IEEE},
          \and
          Zhaoxiang~Zhang,~\IEEEmembership{Senior~Member,~IEEE}

\thanks{This work was partially supported by the National Natural Science Foundation of China (NSFC) under Grant No. 62606500, 42476194 and U24A20242, the Natural Science Foundation of Shandong Province under Grant No. ZR2024MF097, and 2026HWYQ-026, and Natural Science Foundation of Hainan Province under Grant No. 425MS106.}
\thanks{T. Li, Y. Wang, and G. Zhong are with the College of Computer Science and Technology, Faculty of Information Science and Engineering, Ocean University of China, Qingdao 266100, China (e-mail: \{15220071995@163.com, yuxiwang93@gmail.com, gqzhong@ouc.edu.cn\})}.
\thanks{Z. Zhang is with the New Laboratory of Pattern Recognition (NLPR), State Key Laboratory of Multimodal Artificial Intelligence Systems, Institute of Automation, Chinese Academy of Sciences (CASIA), and University of Chinese Academy of Sciences (UCAS), Beijing 100049, China (e-mail: zhaoxiang.zhang@ia.ac.cn).}%
\thanks{Y. Wang and G. Zhong are corresponding authors.}
}


\maketitle

\begin{abstract}
Dynamic 3D Gaussian Splatting (3DGS) has recently emerged as an efficient paradigm for reconstructing dynamic scenes.
Existing approaches predominantly model scene dynamics through deterministic deformation fields, implicitly assuming that each Gaussian follows a unique geometric trajectory over time.
However, this assumption fundamentally breaks down in underwater scenes, where light scattering, refraction, marine snow, and severe visibility degradation introduce substantial physical ambiguity into the imaging process.
To address this problem, we present a physics-guided uncertainty model for underwater dynamic 3DGS (PUGS) that learns to detect the inherent uncertainty induced by underwater image formation and exploits it as a physical signal to guide dynamic scene representation.
Instead of estimating a single deterministic displacement, PUGS parameterizes each Gaussian's motion as a probability distribution characterized by both its expected deformation and physical uncertainty. 
Specifically, we first introduce a stochastic deformation optimization strategy that samples Gaussian trajectories through local reparameterization, encouraging adaptive exploration of multiple physically plausible solutions during training. To progressively consolidate the learned geometry, we further propose an uncertainty stochastic annealing framework together with a dual-path self-consistency distillation objective, which gradually transforms stochastic exploration into deterministic reconstruction while preserving temporal coherence. 
%
%
Extensive experiments on challenging underwater dynamic datasets and physics-aware benchmark demonstrate that the proposed method consistently outperforms state-of-the-art methods in rendering quality, temporal consistency, and reconstruction robustness under complex conditions, achieving an average PSNR improvement of 1.16 dB over the strongest baseline on the UVEB benchmark and 3.99 dB on DRUVA.
Project page is that: \url{https://tingtingpius.github.io/PUGS/}.
\end{abstract}


\section{Introduction}

  \begin{figure*}[t]
    \centering
    \includegraphics[width=\linewidth]{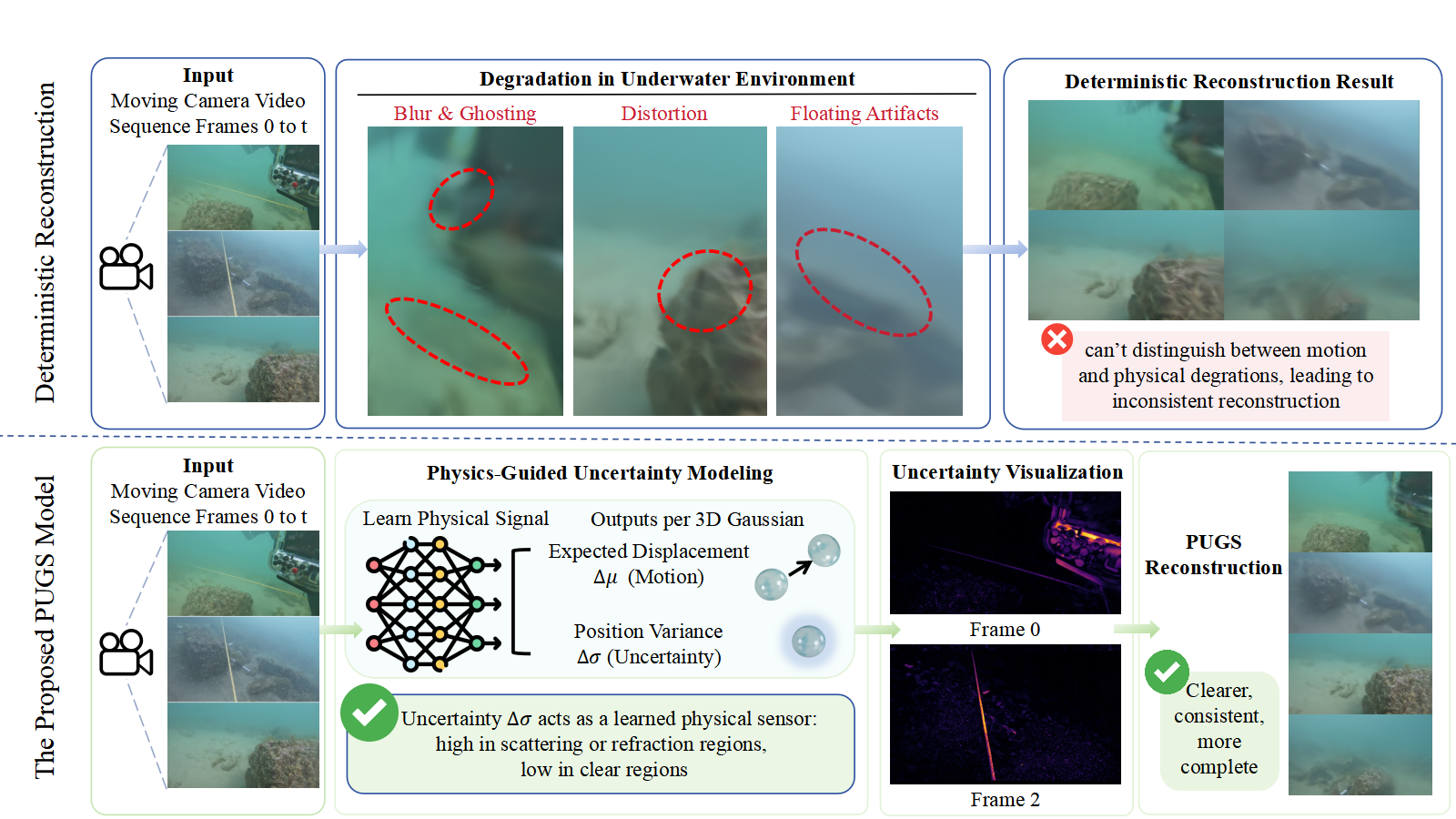}
    \caption{We present PUGS, a method that
  explicitly models physical uncertainty in dynamic 3D Gaussian splatting. On the challenging underwater sequence, a deterministic deformation-based method (Row 1) entangles structural motion with optical interference, producing floaters, blur, distortion, and ghosting artifacts. In contrast, PUGS (Row 2) learns a per-pixel physical uncertainty signal $\Delta\sigma$ (visualized as heatmaps) that captures where and when the optical medium degrades observation reliability. This uncertainty is annealed over time $\beta(t): 1 \rightarrow 0$, enabling the model to progressively resolve physical ambiguity and produce clean, artifact-free renderings that closely match the ground-truth reference.}
    \label{fig:first}
  \end{figure*}

\IEEEPARstart{S}{ynthesizing} photorealistic dynamic environments from monocular footage is a long-standing problem in computer vision and multimedia research. The ability to reconstruct dynamic 3D scenes from a single moving camera underpins a wide range of applications, including autonomous navigation, robotic interaction, and immersive content creation. In particular, underwater scenarios such as an autonomous underwater vehicle (AUV) operation \cite{ma2025state,he2025review,zhong2025cutting}, marine biological monitoring, and virtual oceanography require accurate and temporally consistent 3D reconstructions to enable reliable perception in highly challenging environments. Therefore, advancing dynamic scene reconstruction in underwater settings not only extends the applicability of 3DGS to challenging underwater domains but also carries significant practical implications for ocean exploration. 

The recent emergence of Neural Radiance Fields (NeRF) \cite{mildenhall2020nerf} and 3D Gaussian Splatting (3DGS) \cite{kerbl20233d} fundamentally transforms the landscape of view synthesis. NeRF introduces the concept of representing scenes as continuous volumetric radiance fields optimized via implicit differentiable rendering. However, its high computational cost and slow rendering speed motivate the development of 3D Gaussian Splatting, which represents scenes as sets of explicit 3D Gaussians optimized through rasterization-based rendering, enabling real-time performance while retaining high fidelity. Building upon this, dynamic extensions such as Deformable 3DGS \cite{yang2024deformable} and 4D-GS \cite{wu20244d} are proposed to handle non-rigid motion by learning a deformation field that maps time-varying Gaussians to a canonical space \cite{fan2025advances}. These approaches are demonstrated impressive results on standard dynamic datasets, establishing deformation-based methods as the paradigm for dynamic 3D reconstruction.

Despite their success in terrestrial settings, these state-of-the-art dynamic 3DGS methods face fundamental challenges when deployed underwater. Unlike in-air scenes, underwater environments exhibit severe optical distortions caused by wavelength-dependent absorption, forward and backward scattering, and dynamic refractive distortions \cite{levy2023seathru}. These effects manifest as high-frequency visual artifacts \cite{song2024visual} that are entangled with the actual geometric motion of the scene \cite{zhang2025refractive}. Critically, existing deformation-field models treat every pixel-level change as a deterministic coordinate shift, assuming that all temporal variations correspond to meaningful structural motion. However, in underwater scenarios, marine snow \cite{hodne2022detecting} and refractive caustics introduce stochastic perturbations whose apparent motion is indistinguishable from scene deformation. As shown in Fig. \ref{fig:first}, when a deterministic model attempts to minimize reconstruction loss in these regions \cite{hui2026joint}, this optimization inevitably leads to floating artifacts, temporal inconsistency, and a fundamental failure to distinguish between structural motion and optical interference. Traditional methods \cite{yang2025seasplat,sypek2024waternerf} attempt to address this by deterministically modeling optical properties such as wavelength-dependent attenuation through optical fitting and modifying alpha blending. However, such approaches cannot model the stochastic appearance of marine snow and other particle scattering in underwater environments. It is precisely due to these phenomena that we argue that modeling this uncertainty is essential for robust underwater reconstruction.

These stochastic phenomena share a critical property that is overlooked in prior work. Backscatter intensity correlates with suspended particle density along the line of sight; refractive caustics concentrate at depth-dependent spatial frequencies; marine snow follows characteristic size and velocity distributions~\cite{hodne2022detecting}. In other words, the statistical signature of underwater optical degradation is not arbitrary---it is a \emph{physical signal} that can be detected and exploited. We argue that a model which learns to predict not only \emph{where} each Gaussian moves, but also \emph{how uncertain} that prediction is, effectively learns to sense the local optical properties of the surrounding water medium---without requiring explicit simulation of the forward image formation process.

In this paper, we posit that \textbf{the central challenge of underwater dynamic reconstruction is not merely recovering geometry from degraded observations, but learning to detect and exploit the physical uncertainty signal embedded in the imaging process.} This signal, once extracted, reveals which scene regions are affected by scattering, refraction, or turbidity---enabling the reconstruction to adaptively allocate representational capacity and to disentangle persistent structural motion from transient optical interference. Rather than forcing every observation to correspond to a unique deformation, an ideal representation should distinguish persistent geometric evolution from stochastic optical perturbations, allowing multiple physically plausible hypotheses during optimization before converging to a consistent reconstruction.

Motivated by this observation, we propose a Physics-Guided Uncertainty Model for Underwater Dynamic 3D Gaussian Splatting, called PUGS. Instead of predicting deterministic Gaussian displacements, PUGS reformulates dynamic scene representation from a probabilistic perspective, modeling each Gaussian trajectory as a latent distribution characterized by its expected motion and physical uncertainty. Here, uncertainty is not treated as an auxiliary confidence score, but serves as a compact representation of the stochastic physical effects introduced by underwater image formation. Consequently, the model naturally separates stable structural evolution from transient optical disturbances, yielding significantly more robust deformation estimation in visually degraded environments.
Building upon this probabilistic representation, we further develop a unified uncertainty-aware optimization framework consisting of three complementary components. First, we introduce stochastic Gaussian deformation via local reparameterization, enabling adaptive exploration of multiple physically plausible trajectories during optimization. Second, we propose a dual-path self-consistency distillation mechanism together with uncertainty stochastic annealing, progressively reducing stochasticity while preserving temporal coherence and geometric stability throughout training. Finally, the learned uncertainty is exploited as an explicit reliability cue to guide adaptive Gaussian densification, allocating additional representation capacity to ambiguous regions affected by severe optical degradation or complex non-rigid motion. During inference, uncertainty sampling is disabled, producing deterministic renderings without sacrificing reconstruction fidelity. Extensive experiments on two underwater dynamic and one physical benchmarks demonstrate that PUGS achieves new state-of-the-art performance under severe turbidity and marine snow conditions.

The main contributions of our work are as follows:
\begin{itemize}
\item We identify that deterministic deformation models in dynamic 3DGS fundamentally conflate structural motion with underwater optical perturbations, and propose PUGS---a framework that learns to detect the physical uncertainty signal induced by underwater image formation and exploits this signal to guide dynamic reconstruction.
\item We introduce a stochastic deformation field with reparameterized sampling that predicts per-Gaussian positional uncertainty alongside mean displacements, enabling robust optimization under severe refractive distortions and transient visual noise.
\item We employ a Dual-path Rendering Self-consistency Distillation mechanism that jointly optimizes stochastic and deterministic rendering branches, enforcing long-term structural coherence while enabling local exploration to escape poor minima.
\item We conduct extensive experiments on three dynamic benchmarks---UVEB, DRUVA, and PhysGaia---and demonstrate that PUGS consistently achieves state-of-the-art reconstruction quality across diverse underwater and physical scenes, with comprehensive ablations verifying the contribution of each proposed component.

\end{itemize}

\section{Related Work}

\subsection{Dynamic Scene Reconstruction}
Early neural approaches for dynamic scene reconstruction, such as D-NeRF~\cite{pumarola2021d} and Nerfies~\cite{park2021nerfies}, extended NeRF by learning deformation fields to map time-varying observations to a canonical space, though their purely MLP-based nature incurred significant computational costs. Subsequent hybrid representations like TiNeuVox~\cite{fang2022fast} and HexPlane~\cite{cao2023hexplane} improved efficiency through voxel grids and factorized spatiotemporal volumes. Beyond NeRF-based paradigms, a diverse body of work has explored alternative representations. Traditional methods leverage multi-view stereo and SLAM systems~\cite{taketomi2017visual} to reconstruct dynamic scenes through mesh deformation and keyframe-based optimization. More recently, learning-based approaches emerge, including video segmentation-based techniques that treat dynamic reconstruction as layer decomposition~\cite{tschernezki2021neuraldiff}, and depth-prior methods that incorporate monocular depth estimates to regularize the under-constrained problem~\cite{liang2023semantic}. These methods, while effective in specific domains, often rely on strong priors or suffer from limited generalizability.

With the advent of 3D Gaussian Splatting (3DGS)~\cite{kerbl20233d}, the field shifts toward explicit scene representations that achieve real-time rendering with high fidelity. Dynamic extensions of 3DGS rapidly proliferate. Deformable-3DGS~\cite{yang2024deformable} employs a deformation network to predict time-dependent transformations per Gaussian, while 4D-GS~\cite{wu20244d} adopts HexPlane-based spatiotemporal structures for efficient optimization. Gaussian-Flow~\cite{lin2024gaussian} integrates optical flow supervision, and SC-GS~\cite{huang2024sc} introduces sparse control points to reduce computational overhead. SAGD~\cite{hu2026sagd} further decomposes boundary Gaussians to enforce geometric structure, mitigating the ambiguity of unconstrained 3DGS primitives. Other notable approaches include Spacetime-GS~\cite{li2024spacetime} and 3DGStream~\cite{sun20243dgstream}, which focus on streaming and long-sequence reconstruction. Collectively, these methods\cite{stearns2024dynamic,peng2025street} establish deformation-field-based 3DGS as the prevailing paradigm for dynamic reconstruction. Critically, however, all of these methods inherit a shared assumption: that every observed pixel change originates from geometric motion. In underwater environments, where scattering, refraction, and marine snow produce pixel-level fluctuations that are statistically structured by the physical properties of the water medium, this assumption becomes untenable---the models cannot distinguish a floating particle from a moving surface, because they lack any mechanism to detect the physical signal embedded in these degradations.

Moreover, none of the approaches above incorporate mechanisms to represent or reason about motion uncertainty\cite{guo2026uncertainty}. while methods such as Shape of Motion\cite{wang2025shape} and Mosca\cite{lei2025mosca} introduce new physical prior constraints such as depth maps, 2D tracks, masks, etc. to achieve high-quality motion information and geometric depth estimation, they neglect the reconstruction quality. Our method requires no external supervision signals such as depth maps, optical flow, or 2D tracks---it relies solely on the learned physical uncertainty signal to guide reconstruction. PUGS introduces a variational Bayesian deformation framework that learns to detect this physical signal, enabling robust underwater reconstruction by decoupling structural evolution from stochastic medium interference.

\begin{figure*}[t]
  \centering
  \includegraphics[width=\linewidth]{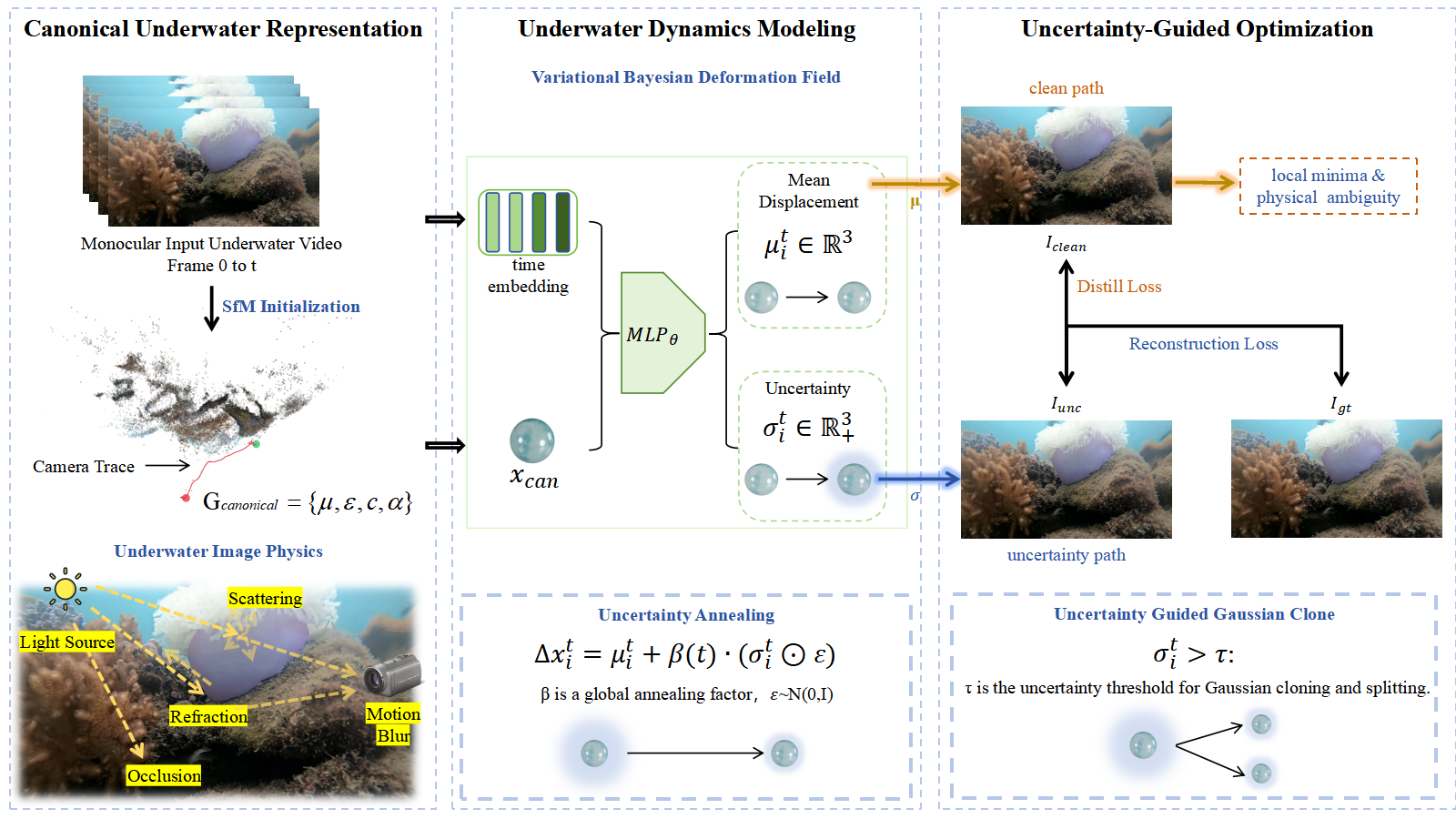} 
  \caption{Overview of the PUGS Pipeline for Underwater Dynamic Scene Reconstruction. Given a monocular video, SfM initializes a set of canonical Gaussians. To address the observational ambiguities caused by underwater physical disturbances, our deformation field predicts not only the displacement mean $\Delta\boldsymbol{\mu}$ but also the positional uncertainty $\Delta\boldsymbol{\sigma}$, yielding a complete characterization of the motion distribution. A deterministic rendering path (clean path) and an uncertainty-aware rendering path (uncertainty path) are then constructed in parallel and jointly optimized via a self-consistency distillation constraint. Meanwhile, Gaussians with high uncertainty are adaptively split to enhance local representation capacity, thereby improving the robustness of underwater reconstruction. In this way, underwater dynamics modeling addresses the failure of deterministic deformation fields under ambiguous observations, while uncertainty-guided optimization leverages the learned uncertainty to allocate representation capacity and stabilize training.}
  \label{fig:pipeline}
\end{figure*}

\subsection{Uncertainty Quantification in Rendering}

The survey by Wang et al.\cite{wang2025review} provides comprehensive coverage of existing methods, but it focuses primarily on uncertainty in 2D vision tasks, where uncertainty-aware pseudo-label refinement~\cite{wang2021uncer} and entropy-based curriculum self-training~\cite{wang2024curriculum} have been extensively studied; in dynamic 3D reconstruction, deformation-level uncertainty remains an open gap. Uncertainty estimation is essential for deploying neural rendering methods in safety-critical applications. Existing uncertainty quantification approaches generally fall into three categories: closed-form likelihood models, Bayesian methods, and ensemble methods. Closed-form likelihood models, exemplified by NeRF-W~\cite{martin2021nerfw} and ActiveNeRF~\cite{pan2022activenerf}, model ray colors using probabilistic distributions to handle transient occluders in static scenes by learning per-pixel variance that re-weights the photometric loss. Bayesian methods such as S-NeRF~\cite{shen2021stochastic} , CF-NeRF~\cite{shen2022conditional} and LP-BNN\cite{franchi2023encoding} learn posterior distributions over radiance fields via variational inference, enabling epistemic uncertainty estimation at the cost of sampling overhead. Ensemble methods, including Density-aware NeRF Ensembles (DANE)~\cite{10161012}, train multiple models independently, providing robust uncertainty estimates but imposing significant computational burdens.

With the emergence of 3D Gaussian Splatting, recent work begins exploring uncertainty quantification within this explicit paradigm. Methods like horseshoe Splatting~\cite{wuhorseshoe,li2024variational}introduce a Bayesian extension of 3DGS with factorized variational inference for calibrated uncertainty estimation\cite{vandemaele2024variational}. RU4D-SLAM~\cite{zhao2026ru4d} incorporates uncertainty-aware perception into Gaussian Splatting SLAM for dynamic scenes through semantic-guided reweighting. More relevant to dynamic reconstruction, FUGS~\cite{shao2025frequency}  and RobustSplat \cite{fu2025robustsplat} explicitly models motion uncertainty by learning uncertainty relations among neighboring Gaussians. Additionally, FisherRF~\cite{jiang2024fisherrf} takes an information-theoretic approach for downstream tasks such as view selection\cite{hanson2025pup}.

However, the aforementioned methods\cite{feng2025gaussian} share a common characteristic: they treat uncertainty as a confidence score---a scalar weight that modulates the contribution of each pixel or Gaussian to the reconstruction loss. While effective for outlier rejection, this passive formulation does not capture where or why the uncertainty arises, nor does it use uncertainty to actively guide the reconstruction process. 
PUGS takes a fundamentally different approach. Rather than treating uncertainty as a post-hoc confidence estimate, we treat it as a physical signal. The position variance $\Delta\boldsymbol{\sigma}$ predicted by our variational Bayesian deformation field is not an abstract confidence measure---it is a learned detector that responds to the statistical signatures of underwater optical degradation. When scattering or refractive distortion dominates a scene region, $\Delta\boldsymbol{\sigma}$ increases, signaling that the model should expand its exploration radius and allocate additional Gaussian capacity. When the water is clear and observations are reliable, $\Delta\boldsymbol{\sigma}$ attenuates, and the model converges to near-deterministic behavior. This active sensing-and-guidance loop---where the physical signal is first detected and then exploited for exploration, refinement, and densification---distinguishes PUGS from all existing uncertainty-aware rendering methods.

\subsection{Underwater Vision and Reconstruction}

Underwater imaging is fundamentally constrained by the physical properties of the aquatic medium \cite{li2024breakthrough,wang2025uwgs}. As described by the Jaffe-McGlamery model \cite{jaffe2002computer}, light propagation underwater is governed by wavelength-dependent absorption and forward/backward scattering, which collectively degrade image quality through color distortion, contrast reduction, and haze-like artifacts\cite{anwar2020diving}. Traditional restoration methods focused on 2D image enhancement, employing hand-crafted priors such as the Dark Channel Prior (DCP) \cite{he2010single} to compensate for these degradations through post-processing techniques. While effective for improving visual quality, these 2D approaches lack any notion of 3D geometry, limiting their applicability to tasks requiring spatial understanding.

The transition to 3D reconstruction in underwater environments is driven by integrating medium-specific physical models into neural rendering frameworks. WaterNeRF~\cite{wang2022waternerf} first extends NeRF to underwater scenes by incorporating attenuation coefficients into the volume rendering pipeline, enabling joint estimation of scene geometry and water parameters from monocular video. SeaThru-NeRF~\cite{levy2023seathru} further advances this direction by modeling both absorption and scattering effects, achieving more accurate color restoration and geometry reconstruction in turbid conditions. These works\cite{zhang2025recgs,liu2025spatiotemporal} demonstrate that implicitly accounting for the underwater medium is essential for high-fidelity 3D reconstruction, yet they remain confined to static scenes.

Extending underwater reconstruction to dynamic 4D scenarios: capturing marine life, moving AUVs, or time-varying water conditions, introduces additional complexity. In dynamic settings, the coupling of fluid dynamics with optical distortions makes it difficult to distinguish genuine scene motion from medium-induced artifacts: a floating particle and a moving fish scale can produce nearly identical pixel-level changes under refractive distortion. Existing dynamic reconstruction frameworks, which assume deterministic motion models, often misinterpret refractive fluctuations and suspended particles as geometric changes, leading to "geometric bleeding" and temporal incoherence. To our knowledge, PUGS is the first framework to address this challenge by learning to detect the physical signal\cite{huang2025rsplatting} within a Gaussian Splatting paradigm, decoupling true structural motion from stochastic medium interference.

\section{Method}

In this section, we present PUGS for high-fidelity dynamic scene reconstruction grounded in a physical uncertainty strategy, with a particular focus on challenging scenarios such as underwater environments. To address the susceptibility of deterministic deformation fields to local minima and "floating" artifacts, we formulate uncertainty-aware physics-based motion modeling within a variational inference paradigm.

As illustrated in Figure~\ref{fig:pipeline}, our method consists of three tightly coupled components:
\noindent\textbf{(1) Uncertainty-Guided Dynamic Scene Representation. }
While maintaining a set of static canonical Gaussians as the geometric scaffold, we introduce a stochastic variational deformation field to model per-frame motion. Instead of predicting a single deterministic displacement, the field simultaneously predicts a mean and a per-Gaussian position variance, treating the variance as a representation of uncertainty and thereby encoding physical uncertainty as a learnable distribution.
\noindent\textbf{ (2)  Uncertainty-Guided Adaptive Refinement with Annealing. }
The learned position variance not only serves as an indicator for triggering adaptive Gaussian splitting and densification in high-uncertainty regions, but is also multiplied by the annealing factor $\beta(t)$, progressively diminishing its own contribution, ensuring that refinement operates on increasingly stable geometry as training converges.
\noindent\textbf{(3)  Dual-Path Rendering with Self-Consistent Distillation. }
During training, we construct a uncertainty-aware rendering path via reparameterization and a parallel deterministic path using only the mean displacement. A self-consistency distillation loss aligns the deterministic prediction with the visual consensus of the uncertainty path, encouraging the deterministic output to capture the central tendency of the explored distribution.

\subsection{Uncertainty-Guided Dynamic Scene Representation }

In underwater dynamic scenes, the observed pixel intensity at the camera is not a deterministic function of 3D geometry alone. Rather, it is the resultant of three interacting optical pathways: direct transmission, forward scattering, and backscatter~\cite{levy2023seathru}. The latter two components introduce stochastic fluctuations whose magnitude depends on the local density of suspended particles, the optical properties of the water column, and the relative geometry among the light source, the scene, and the camera. Critically, these fluctuations exhibit strong spatial structure: regions with high turbidity or dense particle concentrations produce substantially larger frame-to-frame variance than regions in clear water. This makes the \emph{magnitude} of optical degradation a detectable physical quantity.

Our core idea is to learn this quantity. Rather than modeling the forward physics of underwater image formation explicitly, we train a neural deformation field to predict, alongside each Gaussian's expected displacement, a position variance $\Delta\boldsymbol{\sigma}$ that captures the local magnitude of physically induced observational ambiguity. When $\Delta\boldsymbol{\sigma}$ is large, it signals that the corresponding image region is likely affected by scattering, refraction, or particulate occlusion---the model is, in effect, \emph{learning to sense the underwater optical environment} through the statistics of its own reconstruction uncertainty. This learned physical signal subsequently drives two complementary processes: stochastic exploration with progressive annealing and uncertainty-guided Gaussian densification. 

We adopt a canonical-space representation where scene geometry and appearance are defined in a static, time-invariant reference frame. Building upon 3D Gaussian Splatting, each scene primitive is represented by SFM \cite{schonberger2016structure} as an anisotropic 3D Gaussian characterized by its center position $\boldsymbol{\mu} \in \mathbb{R}^3$, covariance matrix $\boldsymbol{\Sigma} \in \mathbb{R}^{3 \times 3}$, opacity $\alpha \in [0,1]$, and view-dependent color $c$ encoded via spherical harmonics. The rendering process projects these Gaussians onto the image plane via differentiable splatting. For a given pixel $\mathbf{p}$, the rendered color $\mathbf{C}(\mathbf{p})$ is computed as
\begin{equation}
    \mathbf{C}(\mathbf{p}) = \sum_{i \in \mathcal{N}} c_i \alpha_i \prod_{j=1}^{i-1}(1-\alpha_j) \mathcal{G}(\mathbf{p}; \boldsymbol{\mu}_i, \boldsymbol{\Sigma}_i),
\end{equation}
where $\mathcal{G}$ denotes the 2D Gaussian kernel obtained by projecting the 3D covariance, and primitives are sorted by depth. To model dynamic scenes, traditional methods introduce a learnable deformation field $\phi $ that only maps canonical positions $\mathbf{x}_{\text{can}}$ to time-dependent deformed positions $\mathbf{x}_{\text{t}}$:
\begin{equation}
    \mathbf{x}_t = \mathbf{x}_{\text{can}} + \phi(\mathbf{x}_{\text{can}}, t).
\end{equation}
This traditional formulation \cite{yang2024deformable} allows the network to learn smooth and plausible motions while maintaining a compact scene representation. However, traditional deformation networks predict deterministic displacements, which inherently struggle with ambiguous scenarios such as rapid motion, optical distortions, or occlusions. These ambiguities manifest as optimization challenges, trapping the learning process in suboptimal local minima and resulting in characteristic artifacts like floating Gaussians or geometric inconsistencies. This issue is especially severe in underwater scenes, where complex physical motions, light refraction, and scattering collectively degrade the imaging process. We address this fundamental limitation by reformulating deformation prediction as a probabilistic process, thereby substantially mitigating these motion ambiguities. 

Our deformation network $\phi_\theta $ takes as input a canonical position $\mathbf{x}_{\text{can}} \in \mathbb{R}^3$ and a time embedding $\mathbf{e}_t \in \mathbb{R}^d$ , and outputs a distribution over deformations rather than a single displacement vector. Specifically, the network predicts both the mean displacement $\Delta\boldsymbol{\mu} \in \mathbb{R}^3$ and position variance $\Delta\boldsymbol{\sigma} \in \mathbb{R}^3$ :
\begin{equation}
    \phi_\theta(\mathbf{x}_{\text{can}}, t) \rightarrow (\Delta\boldsymbol{\mu}, \Delta\boldsymbol{\sigma}).
\end{equation}

The mean displacement $\Delta\boldsymbol{\mu}$ represents the expected motion trajectory, while the variance $\Delta\boldsymbol{\sigma}$ quantifies the physical position uncertainty at each spatial location and temporal instant.
The variance $\Delta\boldsymbol{\sigma}$ can be understood as an adaptive exploration radius. In regions with clear, unambiguous observations, the network learns to predict small variances, leading to deterministic, precise positioning. Conversely, in regions exhibiting physical uncertainty such as motion blur, optical distortions, or occlusions, larger variances permit broader exploration, allowing the optimization to discover geometrically consistent solutions that might be missed by deterministic approaches. 

Our formulation uses an $\ell_1$ uncertainty sparsity loss on $\Delta\boldsymbol{\sigma}$ in place of an explicit KL divergence term. This choice is both pragmatic and physically motivated. A zero-centered Laplace prior (of which $\ell_1$ regularization is the maximum a posteriori counterpart) encodes the inductive bias that physical uncertainty should be sparse while only regions affected by scattering, refraction, or occlusion should exhibit substantial uncertainty. The uncertainty sparsity penalty thus encourages the model to allocate its exploration budget exclusively to regions where the underwater medium has genuinely degraded the visual signal, preventing variance inflation in clear, unambiguous areas.

\subsection{Uncertainty-Guided Adaptive Refinement  with Annealing} 
 
During training, we employ the reparameterization strategy to enable stochastic exploration of the deformation space while maintaining gradient flow. The actual displacement applied to each Gaussian is sampled as :
\begin{equation}
   \Delta\mathbf{x} = \Delta\boldsymbol{\mu} + \beta(t) \cdot (\Delta\boldsymbol{\sigma} \odot \boldsymbol{\epsilon}), \quad \boldsymbol{\epsilon} \sim \mathcal{N}(\mathbf{0}, \mathbf{I}),
\end{equation}
where $\odot$ denotes element-wise multiplication and $\beta(t) \in [0,1]$ is a global stochastic annealing (SA) factor that controls the exploration magnitude. This formulation allows Gaussians to stochastically jitter around their predicted mean positions, effectively exploring the local optimization landscape. The key insight is that this stochastic perturbation enables the optimization to escape sharp local minima\cite{kheradmand20243dgsmcmc} that commonly arise from physically ambiguous observations in challenging underwater scenarios.

The reparameterized sampling in Eq.~(4) can be understood as a physics-aware exploration mechanism. In regions where $\Delta\boldsymbol{\sigma}$ is large the stochastic perturbation automatically expands the exploration radius, enabling the optimization to evaluate multiple physically plausible configurations. In clear, well-observed regions where $\Delta\boldsymbol{\sigma}$ is small, the perturbation naturally shrinks, and the deformation behaves near-deterministically. The annealing factor $\beta(t)$ globally modulates this behavior: early in training, it permits broad exploration to escape poor local minima induced by optical ambiguity; as training converges, it suppresses perturbation, allowing the model to settle into a temporally coherent reconstruction guided by the now-stabilized physical signal.

Furthermore, traditional Gaussian splitting strategies consider only the spatial extent of Gaussian primitives. We extend this criterion by treating the learned physical uncertainty $\Delta\boldsymbol{\sigma}$ as an explicit signal for adaptive capacity allocation. When a Gaussian exhibits position variance exceeding a predefined threshold, it indicates that the underlying scene region is subject to severe underwater optical degradation---dense scattering, refractive caustics, or marine snow occlusion---that cannot be adequately represented at the current spatial resolution. We therefore split such high-uncertainty Gaussians, allocating additional primitives to model the complex, medium-induced appearance variations. This mechanism ensures that representational capacity is automatically concentrated where the physical environment most degrades visual information, rather than being uniformly distributed across the scene.

\subsection{Dual-Path Rendering with Self-Consistency Distillation}

A core contribution of PUGS is the \textbf{Dual-Path Rendering Self-Consistency Distillation} framework, which jointly optimizes stochastic and deterministic rendering branches to enforce long-term structural coherence while enabling local exploration to escape poor minima induced by optical ambiguity. During training, we construct two parallel rendering paths that share the same underlying Gaussian primitives but differ in how deformations are applied: a deterministic clean path using only the predicted mean displacement $\Delta\boldsymbol{\mu}$ to produce $\mathbf{I}_{\text{clean}}$, and a uncertainty path using the reparameterized displacement from Eq.~(4) to produce $\mathbf{I}_{\text{unc}}$\cite{mishchenko2023noise}. By supervising $\mathbf{I}_{\text{unc}}$ against the ground-truth frame, we implicitly encourage the model to converge toward flat minima in the loss landscape---regions where small perturbations do not catastrophically degrade reconstruction quality. The reconstruction objective integrates pixel-wise $\ell_1$ loss with D-SSIM\cite{choi2025efficient} to capture both high-frequency textures and global structures:
\begin{equation}
    \mathcal{L}_{\text{recon}} = (1 - \lambda_{\text{ssim}})\|\mathbf{I}_{\text{unc}} - \mathbf{I}_{\text{gt}}\|_1 + \lambda_{\text{ssim}}\mathcal{L}_{\text{D-SSIM}}(\mathbf{I}_{\text{unc}}, \mathbf{I}_{\text{gt}}).
\end{equation}

A critical challenge is ensuring that the deterministic expectation $\Delta\boldsymbol{\mu}$ remains an accurate surrogate for the entire motion distribution at inference time, when stochastic sampling is disabled. We resolve this by introducing a \textbf{self-consistency distillation loss} that explicitly aligns the clean path output with the visual consensus formed by the uncertainty path:
\begin{equation}
    \mathcal{L}_{\text{distill}} = \|\mathbf{I}_{\text{clean}} - \text{sg}(\mathbf{I}_{\text{unc}})\|_1,
\end{equation}
where $\text{sg}(\cdot)$ denotes the stop-gradient operation. Crucially, gradients flow exclusively through the clean path, treating the stochastically explored rendering as a fixed visual target; if gradients instead propagated through both paths, the optimization would collapse the uncertainty path toward the clean prediction, nullifying the exploration benefit. This design forces the mean displacement to capture the consensus appearance across the distribution of physically plausible deformations. Concretely, when marine snow or refractive caustics create multiple equally valid pixel configurations, the uncertainty path samples across these alternatives, and the distillation loss pulls the clean path toward their central tendency. This mechanism\cite{fan2025momentumgs} enables PUGS to produce sharp, temporally coherent renderings at inference time despite stochastic training perturbations.

 To prevent the deformation field from inflating $\Delta\boldsymbol{\sigma}$ as a slack variable, we impose an $\ell_1$ uncertainty sparsity loss on the uncertainty field, denoted $\mathcal{L}_{\text{unc}} = |\Delta\boldsymbol{\sigma}|_1$. The complete objective is formulated as:
\begin{equation}
    \mathcal{L}_{\text{total}} = \mathcal{L}_{\text{recon}} + \lambda_{\text{distill}} \cdot\mathcal{L}_{\text{distill}}+\lambda_{\text{unc}}\cdot\mathcal{L}_{\text{unc}}.
\end{equation}
We employ a dynamic scheduling for $\lambda_{\text{distill}}$, progressively shifting the optimization priority from broad stochastic exploration to precise deterministic alignment as training converges, thereby ensuring that the clean path---and thus inference-time rendering---inherits the robustness benefits of variational exploration while ultimately satisfying long-term structural coherence across the temporal sequence.

\section{Experiments}
\label{sec:experiments}

\subsection{Experimental Settings}

\noindent\textbf{Datasets} 
We compare PUGS on three different benchmarks, including two underwater datasets: UVEB~\cite{xie2024uveb} and DRUVA~\cite{Varghese_2023_ICCV}, and a new dataset of complex physical scenes PhysGaia~\cite{kim2026physgaia}.
UVEB is a diverse set of complex dynamic underwater scenes. On this dataset, we run experiments on eleven scenes. These scenes collectively encompass diverse reconstruction challenges, including fine-grained texture details (\textit{Boat}), high-speed non-rigid motion (\textit{Snake}), articulated deformations (\textit{Fish}), complex occlusions (\textit{Shoal}), and severe refractive distortions coupled with ``marine snow'' effects (\textit{Coral}), providing a comprehensive testbed for evaluating dynamic reconstruction under realistic underwater optical conditions.
DRUVA is an underwater dataset featuring highly turbid scenes. In this dataset, we run experiments on four scenes: A1, A2, A11, and A12.
PhysGaia is a new dataset of complex physical scenes, and we run experiments on seven scenes: hanok, single flag, pisa, bouncing balls, basin, tube, and single smoke.

For UVEB and DRUVA, each scene is captured by monocular video in underwater environments, with camera poses estimated using COLMAP~\cite{schonberger2016structure} and formatted to follow the D-NeRF data structure. Following the official protocol of Deformable-GS~\cite{yang2024deformable}, we split each scene into training and test sets. For PhysGaia, we follow the benchmark protocol established in PhysGaia~\cite{kim2026physgaia}.

\begin{table*}[!t]
\centering
\caption{Quantitative comparison on UVEB dataset. We compare our method to state-of-the-art approaches, including static 3D-GS, Sea-splat, 3D-4DGS, PSF-4D, Light-GS, RobustGS and Deformable-GS. We report PSNR, SSIM, and LPIPS. \colorbox{bestc}{\,\,\,}, \colorbox{secondc}{\,\,\,}, and \colorbox{thirdc}{\,\,\,} backgrounds denote the best, second-best, and third-best results, respectively.
}
\resizebox{\textwidth}{!}{%
\begin{tabular}{l|ccc|ccc|ccc|ccc}
\toprule
 & \multicolumn{3}{c|}{Boat} & \multicolumn{3}{c|}{Streak} & \multicolumn{3}{c|}{Fish} & \multicolumn{3}{c}{Cave} \\
Method & PSNR$\uparrow$ & SSIM$\uparrow$ & LPIPS$\downarrow$ & PSNR$\uparrow$ & SSIM$\uparrow$ & LPIPS$\downarrow$ & PSNR$\uparrow$ & SSIM$\uparrow$ & LPIPS$\downarrow$ & PSNR$\uparrow$ & SSIM$\uparrow$ & LPIPS$\downarrow$ \\ \midrule
3D-GS &  24.03& 0.829&  0.309& 24.53 & 0.897& 0.096& 24.75& 0.815& 0.230& 28.53& 0.928& 0.344\\
 Sea-splat& 24.87&  0.779& 0.337& 28.72&  0.934&  \second{0.058}& 26.98&  0.822& 0.262& 29.61& 0.932&0.269\\
 3D-4DGS& 24.22& \third{0.854}& 0.306& 29.04& 0.905& 0.063& 32.03& 0.917& 0.130& 24.73& 0.884&0.283\\
 RobustGS& 17.81 & 0.832& 0.233& 22.95& 0.807& 0.153& 25.58& 0.780& 0.189& 26.14& 0.949&0.197\\
 PSF-4D& 25.05& 0.812& 0.320&  \third{32.41}& \third{0.953}& \best{0.044}& 27.13& 0.903& 0.242& \third{35.45}& \second{0.966}&\third{0.139}\\
 Light-GS& \third{25.13}& \second{0.861}& \second{0.195}& 28.53& 0.912& 0.075& \second{33.42}& \best{0.955}& \best{0.109}& 32.25& \third{0.961}&\second{0.114}\\
Deformable-GS& \second{25.77}&     0.843& \third{0.231}& \second{32.49}& \second{0.956}& 0.062& \third{32.87}& \third{0.925}& \third{0.120}& \second{37.27}& 0.947& 0.272\\
Ours & \best{29.17}&   \best{0.875}& \best{0.190}& \best{32.78}& \best{0.959}& \second{0.056}& \best{33.47}& \second{0.927}& \second{0.111}& \best{41.01} & \best{0.995}& \best{0.009}\\ \midrule

& \multicolumn{3}{c|}{Strait fishes} & \multicolumn{3}{c|}{Puffer} & \multicolumn{3}{c|}{Coral} & \multicolumn{3}{c}{Ground} \\
Method & PSNR$\uparrow$ & SSIM$\uparrow$ & LPIPS$\downarrow$ & PSNR$\uparrow$ & SSIM$\uparrow$ & LPIPS$\downarrow$ & PSNR$\uparrow$ & SSIM$\uparrow$ & LPIPS$\downarrow$ & PSNR$\uparrow$ & SSIM$\uparrow$ & LPIPS$\downarrow$ \\ \midrule
3D-GS &  25.79& 0.823&  0.313& 25.54& 0.826& 0.230& 20.64& 0.858& 0.230& 27.05&  0.786& 0.318\\
 Sea-splat& 26.03&  0.822& \third{0.302}& 26.10&   0.871&  0.171& 26.69&  0.910& 0.083& 27.84& 0.833&0.259\\
 3D-4DGS& 19.71& 0.644&  0.572& 16.21& 0.561& 0.715& 28.52& 0.915& \second{0.070}& 24.99& 0.728&0.427\\
 RobustGS& \third{26.72}& 0.850& 0.266& 24.60& 0.857& 0.193& 25.30& 0.893& 0.120& 26.68& 0.815&0.278\\
 PSF-4D& \second{27.03}& \second{0.860}& \second{0.294}& \third{28.16}& 0.886& 0.172&  \best{ 31.93}& 0.910& 0.280& \second{29.50}& \third{0.863}&0.250\\
 Light-GS& 26.65&  \best{0.864}& 0.358& 27.13& \third{0.894}& \third{0.147}& 27.82& \third{0.918}& \third{0.082}& 28.54& 0.849&\third{0.245}\\
Deformable-GS& 26.43&     0.816&  0.302&  \second{29.30}& \best{0.919}& \second{0.142}& \third{29.88}& \second{0.931}& 0.092& \third{29.32}& \second{0.875}& \second{0.224}\\
Ours & \best{27.11}&   \third{0.853}&  \best{0.276}& \best{29.51}& \second{0.918}& \best{0.141}& \second{31.17}& \best{0.951}& \best{ 0.069}& \best{29.98}& \best{0.882}& \best{0.188}\\ \midrule

& \multicolumn{3}{c|}{Snake} & \multicolumn{3}{c|}{Shoal} & \multicolumn{3}{c|}{Anemone} & \multicolumn{3}{c}{Average} \\
Method & PSNR$\uparrow$ & SSIM$\uparrow$ & LPIPS$\downarrow$ & PSNR$\uparrow$ & SSIM$\uparrow$ & LPIPS$\downarrow$ & PSNR$\uparrow$ & SSIM$\uparrow$ & LPIPS$\downarrow$ & PSNR$\uparrow$ & SSIM$\uparrow$ & LPIPS$\downarrow$ \\ \midrule
3D-GS &  21.09& 0.865&  0.309& 22.09& 0.799& 0.308& 29.07& 0.875& 0.352& 24.83&  0.846&  0.276\\
 Sea-splat& 29.06&  0.909& 0.136& 22.17&  0.810&  0.354& 28.34&  0.889& 0.314& 26.95&  0.865&0.231\\
 3D-4DGS& 29.53& 0.914& 0.145& 21.11& 0.753& 0.418& 27.02& 0.848&  0.407&  25.19& 0.811& 0.321\\
 RobustGS& 28.97& 0.927& 0.158& 19.47& 0.716& 0.354& 27.74& 0.892& 0.315& 24.72& 0.848&0.223\\
 PSF-4D& \third{29.92}&  0.929& 0.154& \third{22.52}& 0.782& 0.331& \best{31.75}&  \second{0.911}& 0.287&  \third{29.17}&  0.889& 0.228\\
 Light-GS& 29.86& \third{0.932}& \third{0.114}& 22.35& \third{0.841}& \third{0.281}& 30.08& 0.898&  \second{0.278}&  28.34& \third{0.899}&\second{0.182}\\
Deformable-GS& \second{30.32}&     \second{0.936}&  \second{0.099}&  \second{22.84}& \second{0.852}& \second{0.251}& \third{30.13}& \third{0.908}& \third{0.279}&  \second{29.69}& \second{0.901}& \third{0.188}\\
Ours & \best{30.62}&   \best{0.937}& \best{0.098}& \best{22.90}& \best{0.853}& \best{0.249}&  \second{31.68}& \best{0.919}& \best{0.272}&  \best{30.85}& \best{0.915}& \best{0.151}\\\midrule
\end{tabular}%
}
\label{tab:quantitative_synthetic}
\end{table*}

\begin{figure*}[!t]
  \centering
  \includegraphics[width=\linewidth]{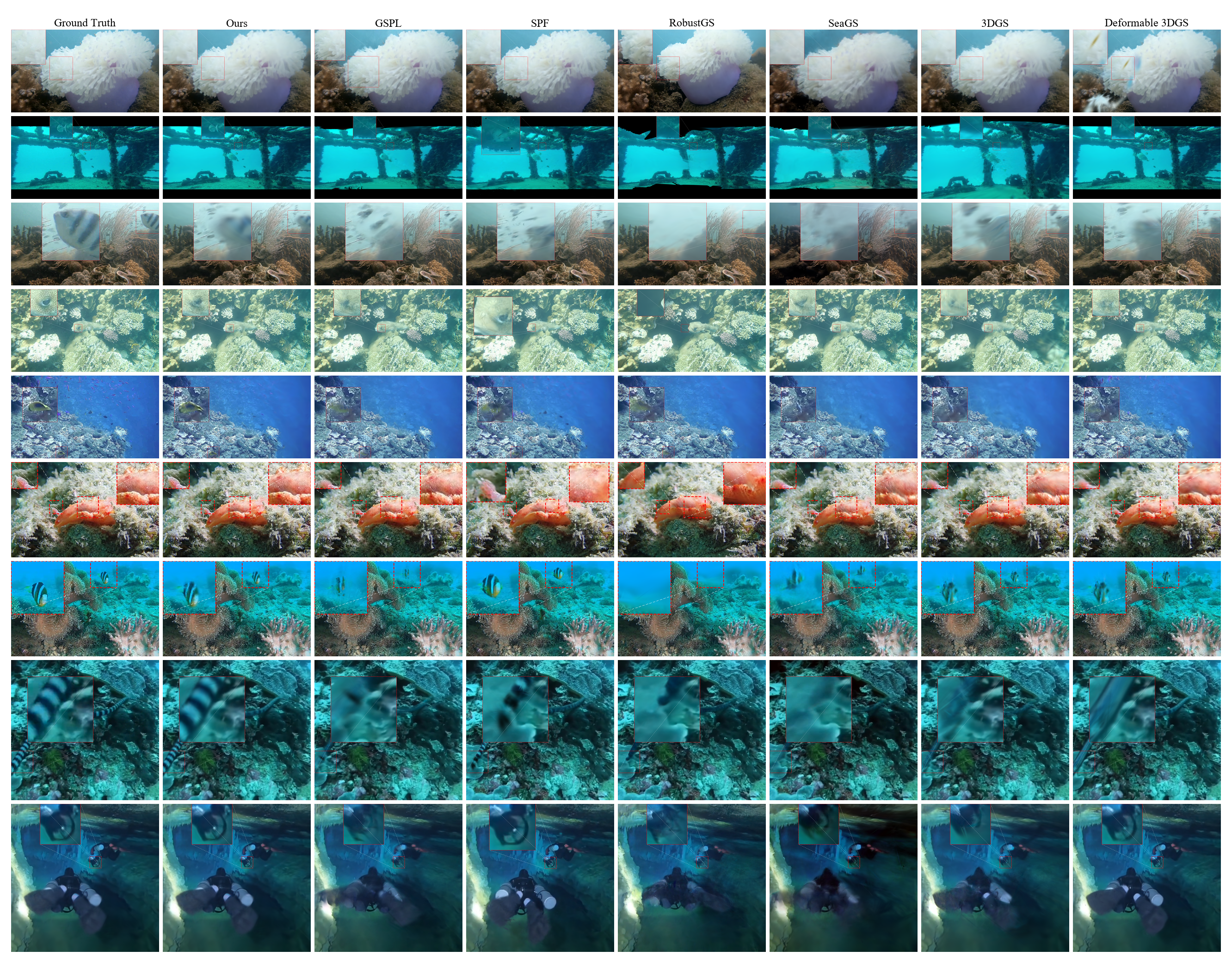} 
  \caption{Qualitative Comparison on UVEB dataset. We conduct comparisons across nine scenes and provide zoomed-in views for each scene to examine the rendered reconstruction details, with a particular focus on dynamic content and complex scene regions. Our method consistently produces higher-quality results in both dynamic and static regions.}
  \label{fig:qualitative}
\end{figure*}

\begin{table*}[!t]
\centering
\caption{Quantitative comparison on DRUVA dataset. We compare our method to state-of-the-art approaches, including TiNeuVox, SeaThru-NeRF, 4DGaussian, 4DGS, WaterSplatting and UDF. We report PSNR, SSIM, and LPIPS.  \colorbox{bestc}{\,\,\,}, \colorbox{secondc}{\,\,\,}, and \colorbox{thirdc}{\,\,\,} backgrounds denote the best, second-best, and third-best results, respectively. 
}
\resizebox{\textwidth}{!}{%
\begin{tabular}{l|ccc|ccc|ccc|ccc|ccc}
\toprule
 & \multicolumn{3}{c|}{A1} & \multicolumn{3}{c|}{A2} & \multicolumn{3}{c|}{A11} & \multicolumn{3}{c|}{A12} & \multicolumn{3}{c}{Average} \\
Method & PSNR$\uparrow$ & SSIM$\uparrow$ & LPIPS$\downarrow$ & PSNR$\uparrow$ & SSIM$\uparrow$ & LPIPS$\downarrow$ & PSNR$\uparrow$ & SSIM$\uparrow$ & LPIPS$\downarrow$ & PSNR$\uparrow$ & SSIM$\uparrow$ & LPIPS$\downarrow$ & PSNR$\uparrow$ & SSIM$\uparrow$ & LPIPS$\downarrow$\\ \midrule
TiNeuVox&  23.05 & 0.757&  0.664& 21.74 & \second{0.843}& 0.637& 27.02 & 0.878& 0.355& 27.59 & 0.911& 0.221& 24.85 & 0.847 & 0.469\\
 SeaThru-NeRF& 26.27 &  0.778& \third{0.336}& 20.23 &  0.694&  0.481& 27.62 &  0.916& 0.254& 26.13 & 0.883 &0.282& 25.06 & 0.818 & 0.338\\
 4DGaussian& 22.84& 0.688& 0.535& 20.93 & 0.586& 0.638& 26.84 & 0.882& 0.273& 25.84 & 0.906&0.244& 24.11 & 0.766 & 0.423\\
 4DGS& 20.75 & 0.736& 0.449& 19.54 & 0.678& 0.516& 24.09 & 0.938& \second{0.173}& 25.79 & \third{0.956}&\second{0.128}& 22.54 & 0.827 & 0.317\\
WaterSplatting& \third{28.71}&     \third{0.830}& 0.376& \third{26.33}& \third{0.809}& \third{0.406}& \third{32.33}& \third{0.962}& \third{0.207}& \third{30.86}& \second{0.958}& \third{0.201}& \third{29.56}& \third{0.890}& \third{0.298}\\
 UDF& \second{29.29}& \second{0.842}& \second{0.300}& \second{27.35}& 0.801& \second{0.384}& \second{34.03}& \second{0.964}&  \best{0.123}& \second{31.01}&  \best{0.969}& \best{0.100}& \second{30.42}& \second{0.894}& \second{0.227}\\
Ours &  \best{33.50}&    \best{0.909}&  \best{0.202}&  \best{31.17}&  \best{0.902}&  \best{0.226}&  \best{37.63}&  \best{0.969}& 0.209&  \best{35.35}& 0.944& 0.233& \best{34.41}& \best{0.931}& \best{0.218}\\ \bottomrule
\end{tabular}%
}
\label{tab:DRUVA_quantitative_synthetic}
\end{table*}

\begin{figure*}[!t]
  \centering
  \includegraphics[width=\linewidth]{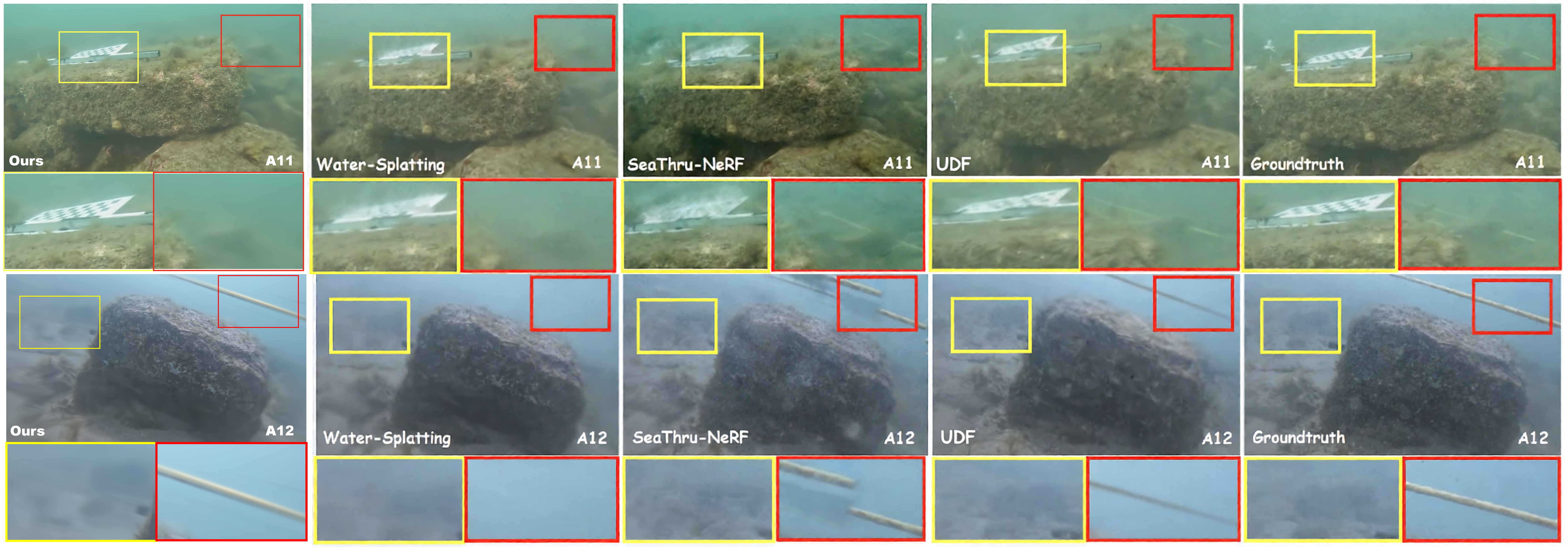} 
  \caption{Qualitative Comparison on DRUVA dataset. We conduct comparisons across two scenes and provide zoomed-in views for each scene to examine the rendered reconstruction details, with a particular focus on dynamic content and complex scene regions. Our method consistently produces higher-quality results in both dynamic and static regions.}
  \label{fig:qualitative2}
\end{figure*}

\begin{table*}[!t]
\centering
\caption{Quantitative comparison on dynamic scenes. We compare our method to MoSca, STG, SOM, 4DGS and Deformable-GS. We report PSNR, SSIM, and LPIPS. \colorbox{bestc}{\,\,\,}, \colorbox{secondc}{\,\,\,}, and \colorbox{thirdc}{\,\,\,} backgrounds denote the best, second-best, and third-best results, respectively. 
}
\resizebox{\textwidth}{!}{%
\begin{tabular}{l|ccc|ccc|ccc|ccc}
\toprule
 & \multicolumn{3}{c|}{hanok} & \multicolumn{3}{c|}{single flag} & \multicolumn{3}{c|}{pisa} & \multicolumn{3}{c}{bouncing balls} \\
Method & PSNR$\uparrow$ & SSIM$\uparrow$ & LPIPS$\downarrow$ & PSNR$\uparrow$ & SSIM$\uparrow$ & LPIPS$\downarrow$ & PSNR$\uparrow$ & SSIM$\uparrow$ & LPIPS$\downarrow$ & PSNR$\uparrow$ & SSIM$\uparrow$ & LPIPS$\downarrow$ \\ \midrule
MoSca&  14.80& 0.590&  0.420& 15.30& 0.560& 0.500& 16.70& 0.540& 0.570& \third{19.90}& \third{0.810}& \second{0.120}\\
 STG& 14.80&  0.600& 0.410& \best{26.00}&  \best{0.870}&  \best{0.140}& \third{20.10}&  \third{0.680}& 0.330& 11.90& 0.510&0.600\\
 SOM& 14.70& 0.670& 0.400& 16.30& 0.600& \second{0.240}& 17.10& 0.650& 0.410& 17.40& 0.790&0.140\\
 4DGS& \third{15.50}& \third{0.740}& \third{0.320}& 18.00& 0.650& 0.290& 19.30& 0.670& \best{0.260}& \best{22.70}& \second{0.860}&\best{0.100}\\
Deformable-GS& \best{16.30}&     \best{0.780}& \best{0.280}& \third{18.30}& \third{0.680}& \second{0.240}& \second{20.70}& \second{0.720}& \best{0.260}& 19.80& 0.770& 0.150\\
Ours & \second{16.10}&   \second{0.760}& \second{0.290}& \second{18.60}& \second{0.690}& \second{0.240}& \best{20.80}& \best{0.730}& \best{0.260}& \second{21.50}& \best{0.910}& \third{0.130}\\ \midrule
 & \multicolumn{3}{c|}{basin} & \multicolumn{3}{c|}{tube} & \multicolumn{3}{c|}{single smoke} & \multicolumn{3}{c}{average} \\
Method & PSNR$\uparrow$ & SSIM$\uparrow$ & LPIPS$\downarrow$ & PSNR$\uparrow$ & SSIM$\uparrow$ & LPIPS$\downarrow$ & PSNR$\uparrow$ & SSIM$\uparrow$ & LPIPS$\downarrow$ & PSNR$\uparrow$ & SSIM$\uparrow$ & LPIPS$\downarrow$ \\ \midrule
MoSca
& \second{18.40}& 0.620& 0.410&  18.30& 0.860&  0.250& 24.80& 0.950&  0.140&  18.30&  0.700& 0.340\\
 STG
& 16.80& 0.660&  0.430& 22.10& 0.920& 0.120& 24.50& \second{0.970}&   0.100&  19.50&   0.740&0.300\\
 SOM
& 16.30& 0.610& 0.430& 22.50& 0.930& 0.150& 23.90& 0.950& 0.180& 18.30&  0.740& 0.280\\
 4DGS
& 18.00& \second{0.680}& \third{0.380}& \best{28.80}& \best{0.960}& \third{0.080}& \third{26.10}& \second{0.970}& \second{0.080}& \third{21.20}& \third{0.790}&\second{0.220}\\
Deformable-GS& \third{18.20}& \third{0.670}& \second{0.360}& \third{27.70}& \best{0.960}& \best{0.050}& \second{26.50}& \second{0.970}& \second{0.080}& \second{21.10}& \second{ 0.790}&  \best{0.200}\\
Ours & \best{19.00}& \best{0.680}& \best{0.330}& \second{28.10}& \best{0.960}& \second{0.060}& \best{28.10}& \best{0.980}& \best{0.070}& \best{21.70}& \best{0.820}&  \best{0.200}\\ \bottomrule
\end{tabular}%
}
\label{tab:physica_quantitative_synthetic}
\end{table*}

\noindent\textbf{Implementation Details} 
All experiments are conducted on a single NVIDIA RTX 3090 GPU with 24GB memory, using PyTorch~\cite{paszke2019pytorch} as the deep learning framework. For a fair comparison, we train all baseline methods using their official implementations with the hyperparameters recommended in their respective papers, and we unify the training iterations to 40,000 across all methods to ensure consistent optimization budgets. Each scene takes approximately two hours to train, including both our method and the baselines.
%
%
%
We note that the original PhysGaia paper reports results with fewer decimal places than our other two benchmarks. To maintain a consistent presentation precision across all three datasets, we pad the published baseline values with trailing zeros where necessary. No values are altered beyond this formatting adjustment.

\noindent\textbf{Baselines} 
We compare PUGS against representative baseline methods on each benchmark, following the evaluation protocol of each corresponding benchmark.

For benchmark UVEB, a diverse set of complex dynamic underwater scenes, we compare our PUGS against six representative baseline methods spanning static and dynamic reconstruction, domain-specific underwater modeling: (1)~\textbf{3D-GS}~\cite{kerbl20233d}: The original 3D GS framework for static scene reconstruction. (2)~\textbf{Sea-splat}~\cite{yang2025seasplat}: A method designed for underwater scene reconstruction that incorporates wavelength-dependent optical modeling. (3)~\textbf{3D-4DGS}~\cite{oh2025hybrid3d4dgaussiansplatting}: A hybrid static-dynamic Gaussian Splatting framework that combines a static 3D representation with a dynamic 4D component to achieve fast reconstruction of dynamic scenes. (4)~\textbf{Deformable-GS}~\cite{yang2024deformable}: A representative deformation-field-based dynamic 3DGS method that learns per-Gaussian deformations via an MLP mapping spatiotemporal coordinates to transformations of a canonical set of Gaussians. (5)~\textbf{Light-GS }(Gaussian-Splatting-Lightning)~\cite{Yuan_2025_ICCV}: A framework that enables fast reconstruction and real-time rendering of urban-scale scenes while maintaining robustness against appearance variations across multi-view captures. (6)~\textbf{PSF-4D}~\cite{iqbal2026psf}: A framework that proposes a novel 4D scene editing framework that exploits different aspects of the diffusion process to ensure temporal and multi-view coherence during editing. (7) RobustGS\cite{ungermann2024robust} enhances 3D Gaussian Splatting’s robustness to dynamic distractors by combining robust loss functions with distractor segmentation masks, achieving better quantitative and qualitative results than vanilla 3D Gaussian Splatting and RobustNeRF in cluttered scenes. 

For benchmark DRUVA, we adopt the baseline configuration established by UDF~\cite{liu2026udf}, which is the most closely related work to ours in both methodology and problem setting. UDF jointly models underwater structure and view-dependent medium effects over time, sharing a similar design philosophy with PUGS in addressing temporally varying underwater degradations. To ensure a fair and controlled comparison, we follow UDF's evaluation protocol and compare against the same set of baselines reported in their paper. We compare our PUGS against six representative baseline methods across four scenes: (1)~\textbf{TiNeuVox}~\cite{fang2022fast}: a radiance field framework that represents scenes with time-aware voxel features. (2)~\textbf{SeaThru-NeRF}~\cite{levy2023seathru}: a new rendering model for NeRFs in scattering media. (3)~\textbf{4DGS}~\cite{wu20244d}: a holistic representation for dynamic scenes. (4)~\textbf{Water-Splatting}~\cite{li2025watersplatting}: a novel approach that fuses volumetric rendering with 3DGS to handle underwater data effectively. (5)~\textbf{4DGaussian}~\cite{yang2024real}: a model that is conceptually simple, consisting of a 4D Gaussian parameterized by anisotropic ellipses. (6)~\textbf{UDF}~\cite{liu2026udf}: a model that jointly represents underwater structure and view-dependent medium over time.

For PhysGaia, we follow the benchmark protocol established in~\cite{kim2026physgaia} and compare against the same set of baselines evaluated in that work. We compare our PUGS against five representative baseline methods in seven scenes: (1)~\textbf{Deformable-GS}~\cite{yang2024deformable}, (2)~\textbf{4DGS}~\cite{wu20244d}, (3)~\textbf{STG}~\cite{li2024spacetime}, (4)~\textbf{Mosca}~\cite{lei2025mosca}, and (5)~\textbf{SOM}~\cite{wang2025shape}.

\subsection{Quantitative and Qualitative Results}
\label{subsec:results}

\noindent\textbf{Results on UVEB} 
Table \ref{tab:quantitative_synthetic} shows quantitative results in UVEB dataset, comparing eleven underwater scenes. Our PUGS consistently outperforms all baselines. Static 3D-GS fails on complex motions (e.g., Snake 21.09 dB). Sea-splat improves over static methods but lags behind dynamic approaches. Deterministic methods (3D-4DGS, Deformable-GS) perform well on structured motions like Streak and Snake but collapse under challenging conditions.
In contrast, PUGS achieves the highest scores across almost all metrics. On Cave, it reaches 41.01 dB (+3.74 over Deformable-GS) and LPIPS 0.009. On Boat, it reaches 29.17 dB (+3.40 over Deformable-GS) . Consistent gains on Snake and Shoal confirm that uncertainty modeling decouples structural motion from stochastic interference.

Figure \ref{fig:qualitative} shows that deterministic baselines produce floaters and motion blur around object boundaries in the dataset UVEB. Our method recovers sharp textures and clean backgrounds, escaping local minima via stochastic exploration. Consistency distillation ensures temporal coherence. These qualitative results align with our quantitative findings, validating that PUGS decouples scene dynamics from environmental noise for high-fidelity underwater reconstruction.

\noindent\textbf{Results on DRUVA} 
Table~\ref{tab:DRUVA_quantitative_synthetic} reports quantitative results on the DRUVA dataset across four underwater scenes. PUGS achieves the best PSNR on all four scenes and the best SSIM on three scenes, demonstrating strong pixel-level reconstruction accuracy. However, we observe a PSNR-LPIPS trade-off in highly turbid scenes. On A11 and A12, PUGS attains substantially higher PSNR (37.63 dB and 35.35 dB) than the next-best method UDF (34.03 dB and 31.01 dB), yet its LPIPS scores (0.209 and 0.233) lag behind UDF (0.123 and 0.100). We attribute this to variance inflation in regions with dense floating particles. The model accounts for turbidity-induced occlusion by increasing the exploration radius, which can introduce subtle blurring that penalizes perceptual metrics while preserving pixel-wise accuracy. We regard mitigating this trade-off as a direction for future work. Despite this limitation, PUGS outperforms all compared methods, including TiNeuVox, SeaThru-NeRF, 4DGaussian, 4DGS, WaterSplatting, and UDF, in overall PSNR and SSIM.

Figure~\ref{fig:qualitative2} presents qualitative results on the DRUVA dataset. In turbid water, our method exhibits some blurriness: in A11, the rope behind the stone slab appears slightly blurred, and in A12, the background behind the stone slab also shows mild blurring. These artifacts are consistent with the variance inflation behavior discussed above. Nevertheless, even in these challenging cases, PUGS preserves sharper foreground structures compared to baselines.

\begin{table*}[!t]
\centering
\begin{minipage}[t]{0.48\textwidth}
\centering
\fontsize{9pt}{11pt}\selectfont
\renewcommand{\arraystretch}{1.2}
\setlength{\tabcolsep}{6pt}
\caption{Results in Snake scene of the contribution of each proposed component: Stochastic Annealing (SA), Uncertainty Sparsity Loss ($\mathcal{L}_{\text{unc}}$), and Self-Consistency Distillation Loss ($\mathcal{L}_{\text{distill}}$). We report PSNR, SSIM, and LPIPS. \colorbox{bestc}{\,\,\,}, \colorbox{secondc}{\,\,\,}, and \colorbox{thirdc}{\,\,\,} backgrounds denote the best, second-best, and third-best results, respectively.}
\label{tab:abs1}
\begin{tabular}{ccc|ccc}
\toprule
\textbf{SA}& \textbf{$\mathcal{L}_{\text{unc}}$}& \textbf{$\mathcal{L}_{\text{distill}}$}& PSNR$\uparrow$ & SSIM$\uparrow$ & LPIPS$\downarrow$ \\
\midrule
            &               &               & 30.09& 0.933& 0.160\\
            &               & \checkmark    & 30.45& 0.935& 0.102\\
            & \checkmark    &               & 30.40& 0.936& 0.099\\
\checkmark  &               &               & 30.31& 0.935& 0.157\\
            & \checkmark    & \checkmark    & 30.34&  0.936& \second{0.099}\\
\checkmark  &               & \checkmark    & \third{30.48}& \second{0.936}& \second{0.099}\\
\checkmark  & \checkmark    &               &  \second{30.53}& \third{0.936}& 0.100\\
\checkmark  & \checkmark    & \checkmark    & \best{30.62}& \best{0.937}& \best{0.098}\\
\bottomrule
\end{tabular}%
\end{minipage}\hfill
\begin{minipage}[t]{0.48\textwidth}
\centering
\fontsize{9pt}{11pt}\selectfont
\renewcommand{\arraystretch}{1.2}
\setlength{\tabcolsep}{6pt}
\caption{Results in Boat scene of the contribution of each proposed component: Stochastic Annealing (SA), Uncertainty Sparsity Loss ($\mathcal{L}_{\text{unc}}$), and Self-Consistency Distillation Loss ($\mathcal{L}_{\text{distill}}$). We report PSNR, SSIM, and LPIPS. \colorbox{bestc}{\,\,\,}, \colorbox{secondc}{\,\,\,}, and \colorbox{thirdc}{\,\,\,} backgrounds denote the best, second-best, and third-best results, respectively.}
\label{tab:abs2}
\begin{tabular}{ccc|ccc}
\toprule
\textbf{SA}& \textbf{$\mathcal{L}_{\text{unc}}$}& \textbf{$\mathcal{L}_{\text{distill}}$}& PSNR$\uparrow$ & SSIM$\uparrow$ & LPIPS$\downarrow$ \\
\midrule
            &               &               & 28.10& 0.858& 0.231\\
            &               & \checkmark    & 28.12& 0.868& 0.233\\
            & \checkmark    &               & 28.38& 0.866& \third{0.195}\\
\checkmark  &               &               & 28.72& 0.872& 0.197\\
            & \checkmark    & \checkmark    & 28.09&  0.865& 0.230\\
\checkmark  &               & \checkmark    & \second{28.95}& \third{0.872}& 0.200\\
\checkmark  & \checkmark    &               &  \third{28.93}& \second{0.874}& \second{0.192}\\
\checkmark  & \checkmark    & \checkmark    & \best{29.17}& \best{0.875}& \best{0.190}\\
\bottomrule
\end{tabular}%
\end{minipage}
\end{table*}

\begin{figure*}[!t]
\centering
\begin{minipage}[t]{0.48\textwidth}
\centering
\includegraphics[width=\linewidth, height=5.6cm, keepaspectratio]{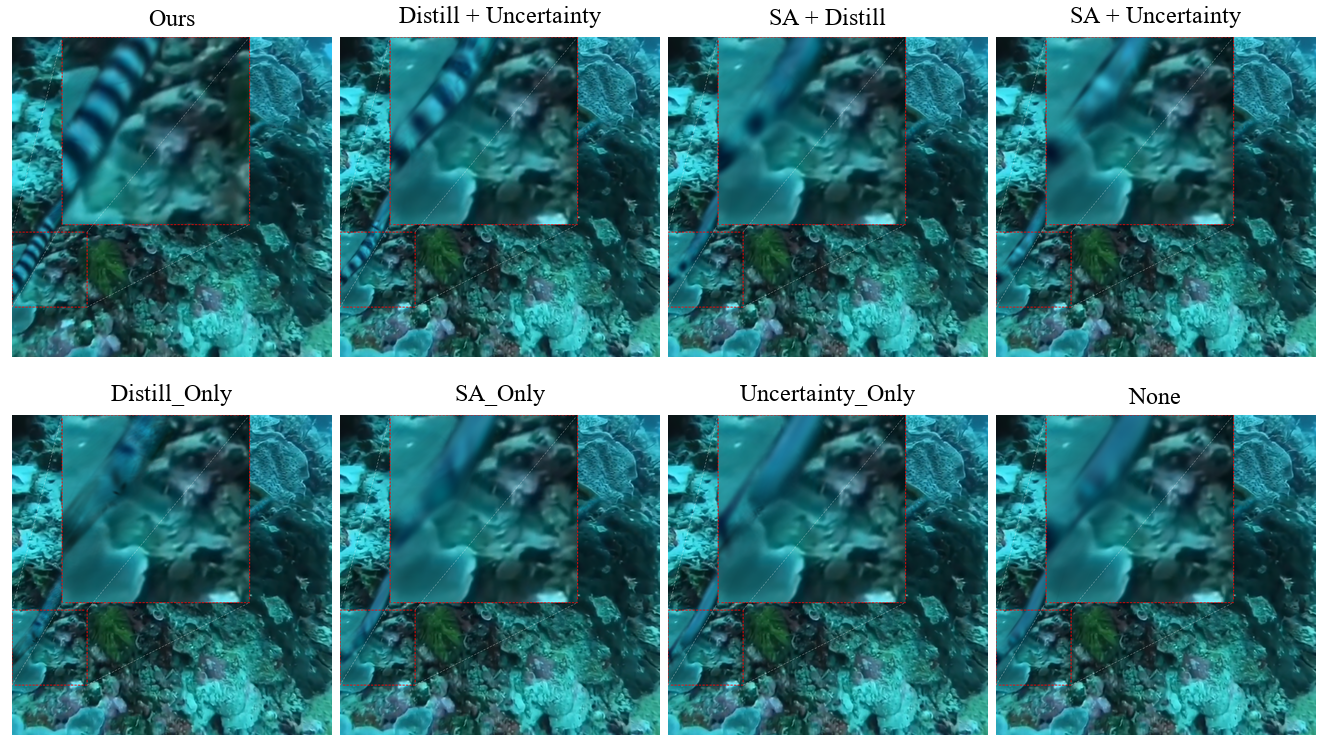}
\caption{We conduct qualitative ablation experiments on the Snake scene, evaluating the contributions of the Stochastic Annealing mechanism, the Distillation Loss, and the Uncertainty Sparsity Loss.}
\label{fig:ab1}
\end{minipage}\hfill
\begin{minipage}[t]{0.48\textwidth}
\centering
\includegraphics[width=\linewidth, height=5.6cm, keepaspectratio]{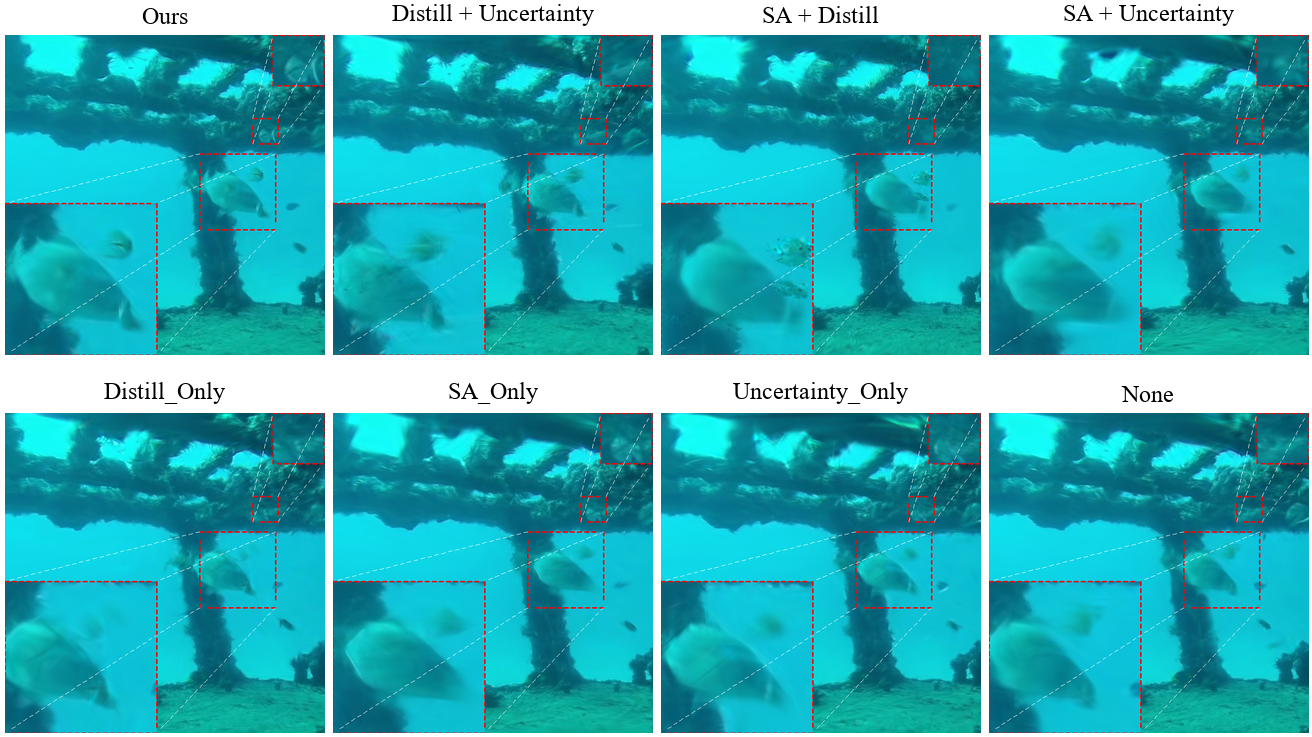}
\caption{We conduct qualitative ablation experiments on the Boat scene, evaluating the contributions of the Stochastic Annealing mechanism, the Distillation Loss, and the Uncertainty Sparsity Loss.}
\label{fig:ab2}
\end{minipage}
\end{figure*}

\noindent\textbf{Results on PhysGaia} 
Table~\ref{tab:physica_quantitative_synthetic} presents the quantitative results on the PhysGaia benchmark. PUGS achieves the best average PSNR of 21.70 dB, outperforming Deformable-GS by 0.60 dB, which demonstrates that the proposed uncertainty-aware framework generalizes beyond underwater domains. A per-scene analysis reveals a pattern consistent with our design philosophy. PUGS delivers its most decisive gains on scenes featuring volumetric, semi-transparent media---\textit{single smoke} (28.10 dB, +1.60 dB over Deformable-GS) and \textit{basin} (19.00 dB, +0.80 dB)---where stochastic appearance variation creates genuine observational ambiguity for the deformation field to resolve. Conversely, on scenes dominated by structured, deterministic dynamics under perfect visibility, such as \textit{hanok} (16.10 dB vs.\ Deformable-GS 16.30 dB) and \textit{single flag} (18.60 dB vs.\ STG 26.00 dB), the stochastic mechanism provides marginal or no benefit, as every pixel-level change corresponds to a unique, resolvable geometric displacement. This divergence confirms that $\Delta\sigma$ genuinely responds to \textit{observational} uncertainty rather than functioning as a generic regularizer, and suggests that a data-driven annealing policy adapting the exploration level to scene statistics could further extend PUGS to broader dynamic scenarios.

\subsection{Ablation Study}

To systematically evaluate the contribution of each algorithmic component and investigate their internal synergies, we conduct a comprehensive ablation study focusing on three core designs of PUGS: the Stochastic Annealing (SA) mechanism, the Uncertainty Sparsity Loss ($\mathcal{L}_{\text{sparse}}$), and the Self-Consistency Distillation Loss ($\mathcal{L}_{\text{distill}}$). Quantitative comparisons across nine configurations in two different scenes are summarized in Table \ref{tab:abs1}, and \ref{tab:abs2}. Ablation experiments are conducted on the Snake scene and Boat scene. 
Qualitative results of the ablation study in Snake scene are presented in Figure \ref{fig:ab1},  qualitative results of the ablation study in Boat scene are presented in Figure \ref{fig:ab2}, further illustrating the visual impact of each component.

\noindent\textbf{(1) Stochastic Annealing Mechanism.}
As shown in Table \ref{tab:abs1}, when SA is introduced alone, the model achieves a significant performance boost, demonstrating that controlled stochasticity enables broader exploration of the solution space, allowing the model to escape local minima induced by refractive noise. However, SA alone yields limited improvement in perceptual quality. This underscores the necessity of coupling stochastic exploration with robust regularization to ensure fine-grained detail recovery. When SA is combined with $\mathcal{L}_{\text{sparse}}$, performance further improves to PSNR 30.53 dB and LPIPS 0.0995, indicating that the sparsity constraint effectively prevents the network from inflating uncertainty arbitrarily while preserving SA's exploratory benefits. Similarly, combining SA with $\mathcal{L}_{\text{distill}}$ yields PSNR 30.48 dB, showing that the distillation loss helps enhance temporal coherence.

\noindent\textbf{(2) Uncertainty Sparsity Loss.}
When applied independently to the baseline, $\mathcal{L}_{\text{sparse}}$ yields a moderate improvement in geometric compactness. This confirms that $\mathcal{L}_{\text{sparse}}$ effectively prevents the network from inflating uncertainty arbitrarily, promoting compact geometry and eliminating floating artifacts. When combined with $\mathcal{L}_{\text{distill}}$, the model achieves PSNR 30.34 dB and LPIPS 0.0990, demonstrating that these two regularization terms work complementarily to enforce both geometric compactness and temporal stability. The best combination involving $\mathcal{L}_{\text{sparse}}$ appears when paired with SA, where the exploratory capability of SA and the compactness constraint jointly contribute to reconstruction.

\noindent\textbf{(3) Self-Consistency Distillation Loss.}
Incorporating $\mathcal{L}_{\text{distill}}$ independently forces the mean prediction to remain robust against stochastic perturbations, leading to substantial improvements in perceptual quality. Visually, this manifests as significantly sharper textures and reduced motion blur. When combined with $\mathcal{L}_{\text{sparse}}$, the two regularization terms together achieve strong performance (PSNR 30.34 dB), demonstrating their complementary roles in enforcing both temporal coherence and geometric compactness.

\definecolor{lightgray}{gray}{0.9} 

\section{Conclusion}\label{sec:conclusion}

In this work, we presented PUGS, a framework that reformulates underwater dynamic scene reconstruction as probabilistic inference over Gaussian trajectories. Unlike deterministic deformation fields that conflate structural motion with optical perturbations, PUGS explicitly models physical uncertainty as a learnable distribution over per-Gaussian displacements. A variational deformation field jointly predicts motion means and position variances, enabling reparameterized stochastic sampling that escapes local minima induced by physically ambiguous observations. Dual-path rendering couples a stochastic exploratory branch with a deterministic inference branch via self-consistency distillation and progressive uncertainty annealing, while the learned variance guides adaptive Gaussian densification toward optically degraded regions. Across three benchmarks spanning 22 dynamic scenes, PUGS achieves state-of-the-art performance, including 30.85~dB PSNR on UVEB (+1.16~dB over Deformable-GS) with pronounced gains of +3.74~dB and +3.40~dB on Cave and Boat, and 34.41~dB on DRUVA (+3.99~dB over UDF). \textbf{Limitations} include a PSNR--LPIPS trade-off under extreme turbidity and the reliance on manually scheduled annealing, both of which motivate future work on data-driven uncertainty scheduling. Beyond underwater domains, this principled integration of physical image formation and uncertainty modeling offers a promising direction for robust reconstruction in fog, smoke, and other scattering media.


\newpage

\begin{IEEEbiography}[{\includegraphics[width=1in,height=1.25in,clip,keepaspectratio]{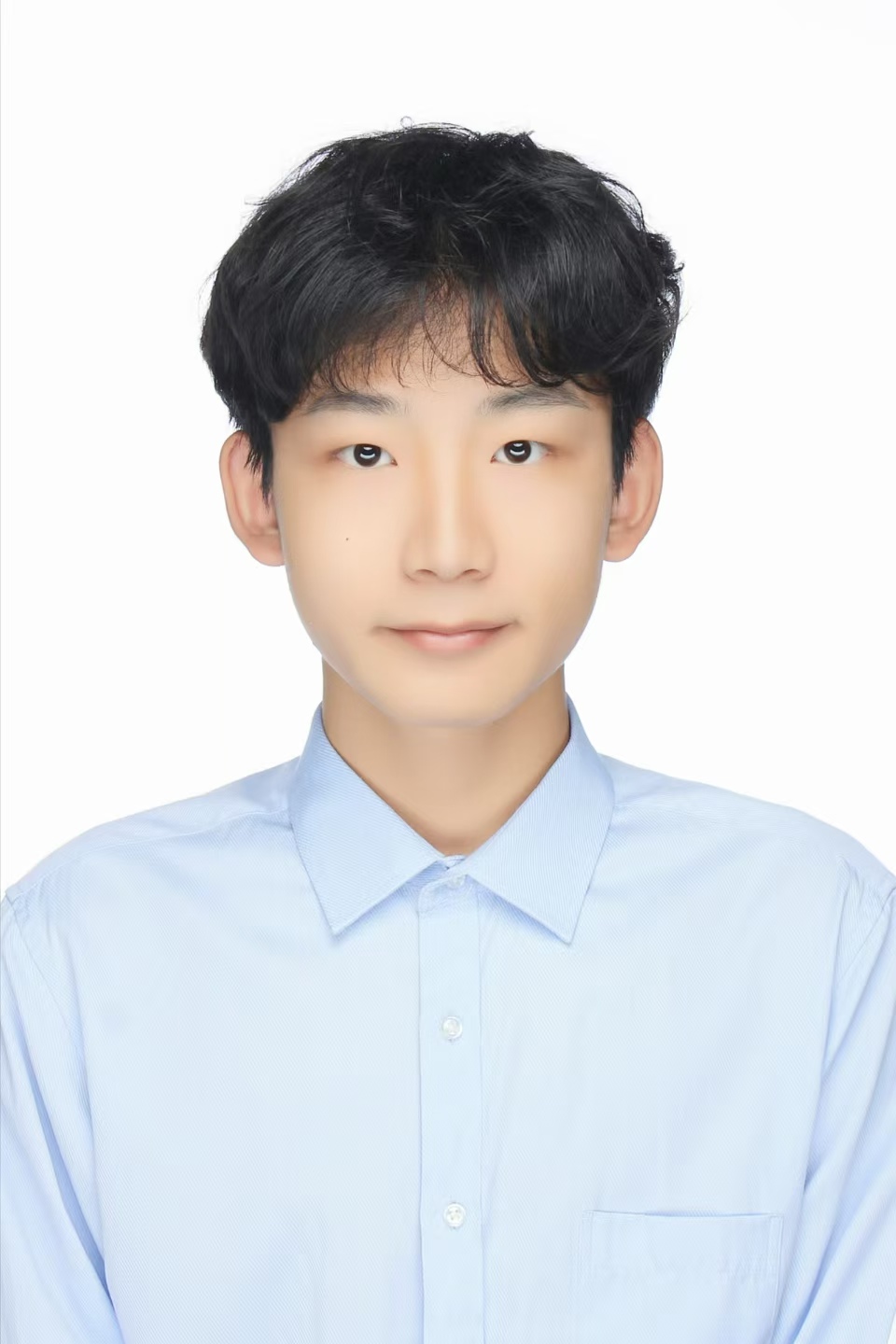}}]
  {Tingjun Li} is currently pursuing his M.E. degree in Computer Science and Technology with the College of Computer Science and Technology, Ocean University of China, Qingdao, China. He received his B.E. degree in Computer Science and Technology from Harbin Institute of Technology, Shenzhen, China, in 2020. His research interests include underwater 3D reconstruction, dynamic neural rendering, and computer vision.
  \end{IEEEbiography}

  \begin{IEEEbiography}[{\includegraphics[width=1in,height=1.25in,clip,keepaspectratio]{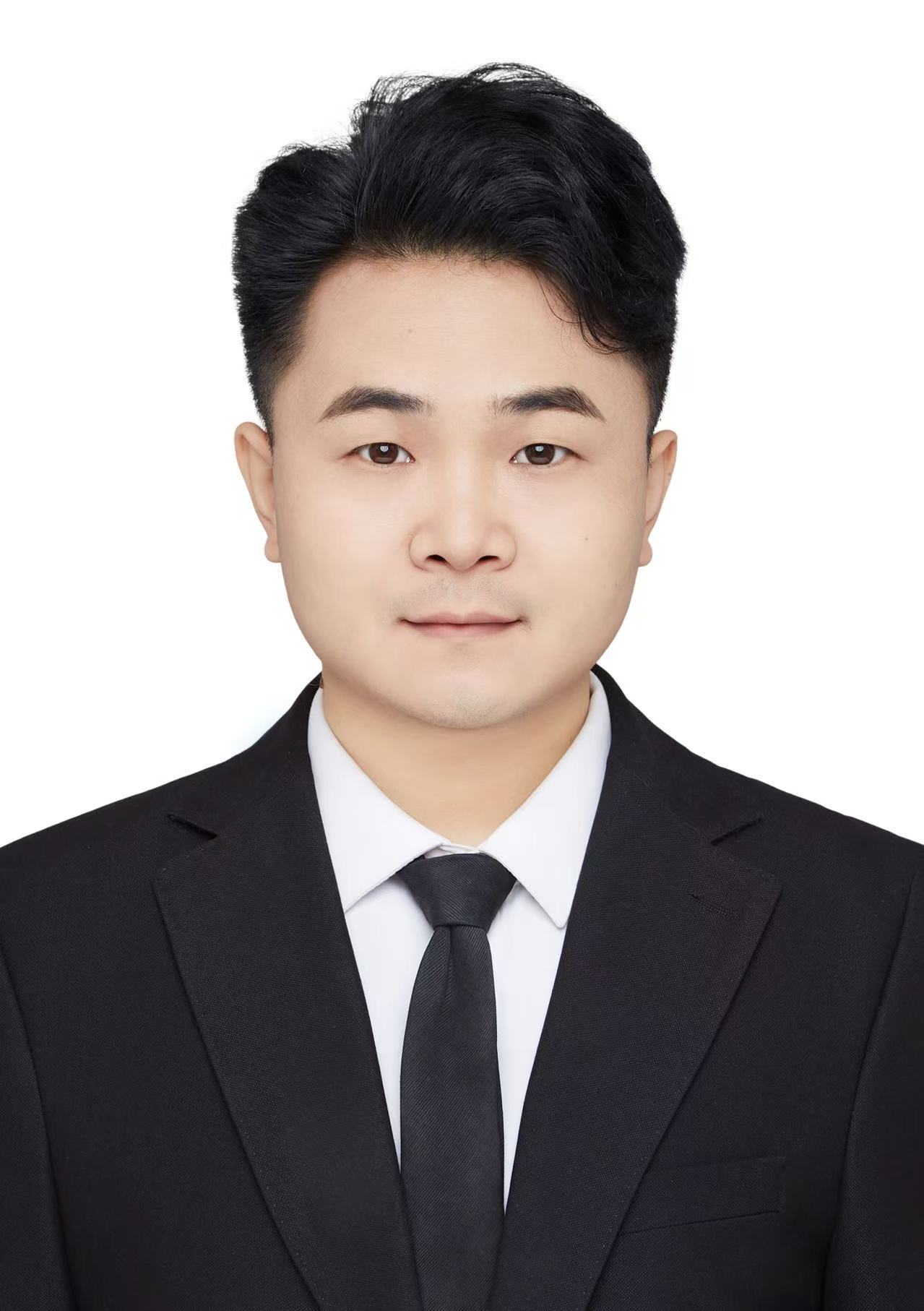}}] {Yuxi Wang} received his Bachelor's degree from Northeastern University, China, in 2016, and his PhD degree from University of Chinese Academy of Sciences (UCAS), Institute of Automation, Chinese Academy of Sciences (CASIA), in January 2022. He is currently an associate professor in the School of Information Science and Engineering, Ocean University of China, Qingdao, 266003, China. Before that, he served as an assistant professor in the Centre for Artificial Intelligence and Robotics (CAIR), Hong Kong Institute of Science \& Innovation, Chinese Academy of Science (HKISI-CAS) from 2022 to 2025. His research interests include 3D AIGC, embodied AI, visual-language navigation, and domain adaptation. He is serving as the Associate Editor of Pattern Recognition. He has also served as the PC of international conferences like CVPR, NeurIPS, ICCV, AAAI, IJCAI and ACM MM.
\end{IEEEbiography}

  \begin{IEEEbiography}[{\includegraphics[width=1in,height=1.25in,clip,keepaspectratio]{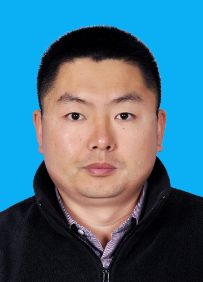}}] {Guoqiang Zhong} received his B.S. degree in Mathematics from Hebei Normal University, Shijiazhuang, China, his M.S. degree in Operations Research and Cybernetics from Beijing University of Technology (BJUT), Beijing, China, and his Ph.D degree in Pattern Recognition and Intelligent Systems from Institute of Automation, Chinese Academy of Sciences (CASIA), Beijing, China, in 2004, 2007 and 2011, respectively. Between October 2011 and July 2013, he was a Postdoctoral Fellow with the Synchromedia Laboratory for Multimedia Communication in Telepresence, University of Quebec, Montreal, Canada. Between March 2014 and December 2020, he was an Associate Professor at Department of Computer Science and Technology, Ocean University of China, Qingdao, China. Since January 2021, he has been a Full Professor at College of Computer Science and Technology, Ocean University of China. His research interests include Pattern Recognition, Machine Learning and Computer Vision.
\end{IEEEbiography}

  \begin{IEEEbiography}[{\includegraphics[width=1in,height=1.25in,clip,keepaspectratio]{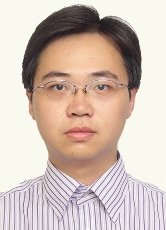}}]{\textbf{Zhaoxiang Zhang}} received his bachelor’s degree from the Department of Electronic Science and Technology in the University of Science and Technology of China (USTC) in 2004. After that, he was a Ph.D. candidate under the supervision of Professor Tieniu Tan in the National Laboratory of Pattern Recognition, Institute of Automation, Chinese Academy of Sciences, where he received his Ph.D. degree in 2009. In October 2009, he joined the School of Computer Science and Engineering, Beihang University, as an Assistant Professor (2009-2011), an Associate Professor (2012-2015) and the Vise-Director of the Department of Computer application technology (2014-2015). In July 2015, he returned to the Institute of Automation, Chinese Academy of Sciences. He is now a full Professor in the Center for Research on Intelligent Perception and Computing (CRIPAC) and the State Key Laboratory of Multimodal Artificial Intelligence Systems.
  His research interests include Computer Vision, Pattern Recognition, and Machine Learning. Recently, he specifically focuses on deep learning models, biologically-inspired visual computing and human-like learning, and their applications on human analysis and scene understanding.
  He has published more than 200 papers in international journals and conferences, including reputable international journals such as JMLR, IEEE TIP, IEEE TNN, IEEE TCSVT, IEEE TIFS and top level international conferences like CVPR, ICCV, NIPS, ECCV, AAAI, IJCAI and ACM MM.
  He is serving or has served as the Associate Editor of IEEE T-CSVT, Patten Recognition, Neurocomputing, and Frontiers of Computer Science. He has served as the Area Chair, Senior PC of international conferences like CVPR, NIPS, ICML, AAAI, IJCAI and ACM MM. He is a Senior Member of IEEE, a Distinguished Member of CCF, and a Distinguished member of CAAI.
\end{IEEEbiography}

\end{document}